\documentclass[11pt,a4paper]{article}
\usepackage[hyperref]{acl2020}
\usepackage{times}
\usepackage{latexsym}

\usepackage{microtype}

\aclfinalcopy 

\usepackage{amsfonts}
\usepackage{amsmath}
\usepackage{graphicx}

\usepackage{color}

\usepackage{enumitem}

\title{Targeting the Benchmark: On Methodology in Current\\ Natural Language Processing Research}

\author{David Schlangen\\
  Computational Linguistics / Department of Linguistics\\
  University of Potsdam, Germany\\
  {\tt david.schlangen@uni-potsdam.de} \\}

\date{}

\begin{document}
\maketitle
\begin{abstract}
It has become a common pattern in our field: One group introduces a \emph{language task}, exemplified by a \emph{dataset}, which they argue is \emph{challenging} enough to serve as a \emph{benchmark}. They also provide a baseline \emph{model} for it, which then soon is \emph{improved} upon by other groups. Often, research efforts then move on, and the pattern repeats itself. What is typically left implicit is the argumentation for why this constitutes progress, and progress towards what. In this paper, we try to step back for a moment from this pattern and work out possible argumentations and their parts.
\end{abstract}

\section{Introduction}
\label{sec:intro}

The goal of any field of research is to make progress towards answering its foundational questions. To do so, a \emph{methodology} is required that guides attempts at providing or improving answer proposals. In natural language processing, the object of study is human language, and any methodology for doing research in this field will need to have some contact with examples of this object. This contact has become more and more direct in the past decades, with samples of language becoming more directly the material from which proposals (in the form of statistical \emph{models}) are derived. Recent years have seen an increase in the collection of samples specifically for the purpose of creating \emph{benchmarks}, against which progress in devising models can be measured. It is this function of \emph{benchmarking}, and its role in a progress-oriented methodology, that this paper aims to investigate.

\begin{figure}[t]
  \centering
  \vspace*{-3ex}
  \hspace*{-3ex}
  \includegraphics[width=1.1\linewidth]{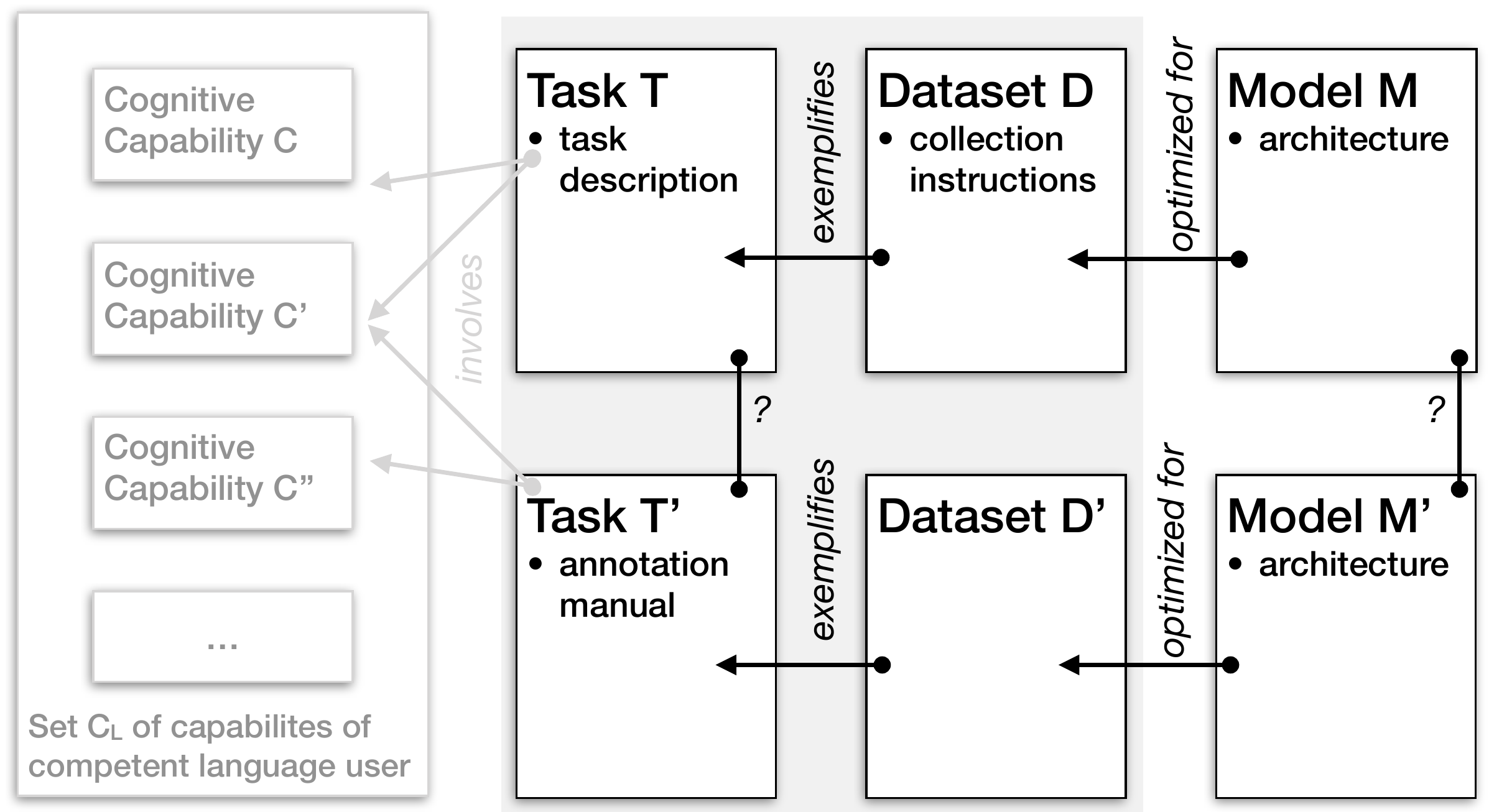}
  \caption{Relations between Research Objects in a Benchmark-Driven Methodology}
  \label{fig:bench}
  \vspace*{-2.5ex}
\end{figure}

Figure~\ref{fig:bench} illustrates the basic structure of a benchmarking methodology: A \emph{language task} is devised that is a) restricted enough to be managable with current methods, and b) deemed challenging for the \emph{capabilities} that it involves.\footnote{This figure is from \cite{schlangen:tasks}, of which this is a shorter version developed in a slightly different direction.}
For this task, a \emph{dataset} is collected, often via crowd sourcing, on which in turn models are trained and compared, using evaluation metrics defined together with the task.
What can we learn by following such a methodology? Let's look at the components first and then at ways in which this methodology is, might, and perhaps should be used.

\section{What is a Language Benchmark?}
\label{sec:lbench}
\vspace*{-1ex}

\subsection{What is a Benchmark?}
\label{sec:bench}

In computing, a benchmark is \emph{``a problem that has been designed to evaluate the performance of a system [which] is subjected to a known workload and the performance of the system against this workload is measured. Typically the purpose is to compare the measured performance with that of other systems that have been subject to the same benchmark test.''} \cite{oxlexcompsci}.

The use of this term in NLP is related: 
here, benchmark tasks are also specifically designed for evaluation; however, an important difference is that what is being evaluated is not a full system that has a separate main purpose, but rather an \emph{algorithm} that is instantiated on the benchmark itself. We will discuss the consequences of this below.

This kind of evaluation of learning algorithm has a long tradition in the field of machine learning research.\footnote{%
  For example, the UCI Machine Learning Repository has been collecting and providing datasets for more than 20 years now \cite{Dua:2019}.
}
In this field, a new algorithm would normally be tested on a large collection of datasets, possibly ranging from classifications of flowers to classification of credit records, with no assumption of any internal connection between the datasets. Again, NLP is different here, as all datasets represent facets of the same underlying phenomenon, language use.

We will argue that these two differences (life outside of benchmarking, and internal connection between tasks) are important, but understudied. But first we look at the notion of a \emph{language task} in some more detail.

\subsection{What is a Language Task?}
\label{sec:lt}

A \emph{language task} is a mapping between an \emph{input space} and an \emph{output} or \emph{action space}, at least one of which contains natural language expressions. The mapping has to conform to a \emph{task description}, which is typically given only informally, making reference to theoretical or pre-theoretical constructs external to the definition, such as ``translation'' or ``is true of''. We call this an \emph{intensional description}. Typically, a task will also be specified \emph{extensionally} through the provision of a \emph{dataset} of examples of the mapping (that is, pairs of state and action).
To collect such a dataset, the task description (e.g., ``classification of entailment relations between sentence pairs'') must be operationalised into a collection instruction (``please mark whether the a situation that is well described by sentence A could normally also be described by sentence B'').

\section{How Can It be Evaluated?} 
\label{sec:mot}

\subsection{Relation Task / Dataset}
\label{sec:tdat}

Given a task and a dataset, the first question to ask is how well the latter exemplifies the former. Investigating this is relatively straightforward.
First, the dataset should be \emph{verified}, which is to check whether the provided input/output pairs can indeed be judged correct relative to the task (in its intensional description). If the examples are collected specifically for the purpose of exemplifying the task, this is the process of controlling annotation, and standard methodologies exist \cite{ArtPoesio:kappa,pustestubbs:annotation}. Care needs to be taken that the task is actually well-defined enough to pose an unambiguous challenge to capable language users.\footnote{%
  \newcite{Pavlick2019}, for example, show that the task of annotation textual entailments can lead to faultless annotator disagreements.
}

\emph{Validating} a dataset is a less formalised process. It comprises arguing that the dataset indeed exemplifies the task intension well. For example, pairs only of images of giraffes and sentences describing them would arguably not exemplify the general task of \emph{image description} very well (even if the descriptions are accurate), while perhaps exemplifying the task of \emph{giraffe image description}.

Another way to evaluate a dataset is by trying to model it. If a model can ``solve'' the dataset even when deprived of information that for theoretical or pre-theoretical reasons is seen to be crucial, the dataset can be considered an unsatisfactory exemplification of the task. E.g., in a \emph{visual (polar) question answering} setting \cite{VQA2015}, if in a dataset all and only the expressions that mention giraffes are true, a model could seize on this fact and perform well without needing the images,
which would be evidence that the dataset is deficient relative to the task description.\footnote{%
  The task of visual question answering provides an interesting example case of such a development. After \citet{VQA2015} introduced the first large scale dataset for this task, it quickly became clear that this dataset could be handled competitively by models that were deprived of visual input \cite[``language bias'', as noted e.g.\ by][]{Jabri2016}. This problem was then addressed by \citet{Goyal2017} with the construction of a less biased (and hence more valid) corpus for the same task.
}

\subsection{Relation Cognitive Capability / Task}
\label{sec:ctask}

While the dataset forms the visible surface of the task, it is the task itself that needs to provide value. We can categorise tasks by how they are embedded in further uses: a \emph{product task} task is one that can be argued to have direct value to consumers (such as translation, or search); an \emph{annotation task} is one where the task description is theoretically motivated and the output a linguistically motivated object; a \emph{benchmark task} -- which is the type that concerns us here -- finally is one which gets its value from how well it tests a particular ability (and nothing else) and how well it discriminates learners based on this ability.\footnote{%
  \newcite{aibench}, analysing AI benchmarks in general, distinguish between \emph{difficulty} (which determines the ability level which must be reached to perform better than chance on a task) and \emph{discrimination} (the slope of the graph ability level vs.\ probability of correct response.
}

For a language benchmark task, the argument roughly goes as follows (even if typically only made implicitly): To be good at task $T$, an agent must possess a set $C_T$ of capabilities (of representational or computational nature). If the $c \in C_T$ are capabilities that competent language users 
can be shown or argued to possess and make use of in using language---let's call the set of these capabilities of a competent language user $C_L$, so that $C_T \subseteq C_L$--- then being able to model these capabilities (via modelling the task) results in progress towards the ultimate goal, which is to model competent language use. And hence, any task $T$ that comes with an interesting set $C_T$ is a good task.\footnote{%
  To give some examples of informal versions of this argument, and chosing papers more or less randomly, here are some quotes:

  From the paper that introduced the \emph{visual question answering} task \cite{VQA2015}: ``What makes for a compelling “AI-complete” task? We believe that in order to spawn the next generation of AI algorithms, an ideal task should (i) require multi-modal knowledge beyond a single sub-domain (such as CV) and (ii) have a well-defined quantitative evaluation metric to track progress. [\dots] Open-ended questions require a potentially vast set of AI capabilities to answer -- fine-grained recognition
  [\dots],
  object detection   [\dots],
  activity recognition   [\dots],
  knowledge based reasoning   [\dots],
  and commonsense reasoning
    [\dots].''


    \citet{Williams2018}, on 
    computing entailments:
    ``The task of natural language inference (NLI)
is well positioned to serve as a benchmark task for research on NLU. [\dots] In particular, a model must handle phenomena like lexical entailment, quantification, coreference, tense, belief, modality, and lexical and syntactic ambiguity.'' }

Under what conditions does this argument work? First of all, the assumed connection to the set of capabilities must indeed be there. We have already seen a way to challenge a claimed connection, through providing a model that can ``solve'' a given task (via a dataset) while not having access to information that, given our analysis of the task and interest in $C_T$, should be involved in the capability.\footnote{%
  Such an attack challenges the claim of there being a \emph{necessary} connection between handling $T$ and possessing capability $c$. It might still very well be that humans can only perform this task if they possess capability $c$ (and all the knowledge involved in it), because they wouldn't be able to pick up the statistical correlations that could be exploited.
}
(Although this challenge in the first instance only targets the dataset and not the task itself.)

Secondly, following usual scientific methodology \cite{popper:logik}, we can rank the value of an instantiation of this argument by how precisely the capability is specified, from the trivially correct ``task T involves the capability to do task T'' to a statement that could be wrong, e.g.\ ``task T involves the capability to compute the syntactic structure of a natural language sentence''. Such a statement must make reference to theoretical constructs belonging to the analysis of cognitive capabilities.

Furthermore, we can rank the motivation given for a task by how explicit it is in delineating the set of capabilities it involves. For a given $c \in C_T$, is ``$c$ as required by $T$'' fully \emph{separable} from any other tasks involving $c$?
Or is ``$c$ as required by $T$'' perhaps all that there is to know about $c$, that is, is $c$ \emph{exhaustively} represented by $T$?

Finally, underlying the benchmarking methodology --- where the benchmark is not just a measuring tool, but also a modelling target --- there has to to be the assumption that some sort of transferable knowledge is generated by modelling $T$, so that what the model has learned about (a sufficiently generally specified) $c$ can be used in other tasks that involve $c$. (Let's call this \emph{transferability}; which strictly speaking is a property of models, not of tasks.) More on this below.

To sum up, a benchmark task gets its value from its connection to a particular facet of language, a particular capability of language users; this in turn 
seems to be difficult to specify without access to terms from theories of the domain, which allow us to name these capabilities.\footnote{%
  And one will indeed find that papers introducing such tasks make mention of terms like \emph{syntax}, \emph{semantics}, \emph{compositionality}, \emph{quantifiers}, etc.
}$^,$\footnote{%
  We can also note that with this focus normally comes a certain top-down approach, where the collected data is not investigated for how exactly the human participants actually solved their task. (But see \cite{emielphdthesis} for a detailed study along those lines, for the task of image description.)
}

\section{How are Language Benchmarks Used?}
\label{sec:use}

In the way that these tasks are set up, as single-step tasks that humans can quickly do (``describe this image'', ``is the elephant [in this image] sleeping?'', ``does sentence A follow from sentence B?''), it is tempting to see a similarity to tasks used in intelligence testing (see e.g. \newcite{borsboom:measuring} for an introduction). There is a crucial difference, however: Where intelligence testing works more in the way standard computing benchmarking works (subjecting the \emph{otherwise functioning} learner to a standardised workload), in NLP, benchmarks are both the testing instrument as well as the training material.\footnote{%
  Not unlike a school that aims to improve its test scores by preparing its students specifically for the tests; in that case, however, this practice would be seen as undermining the value of the test.
}$^,$
\footnote{%
  For a recent paper also discussing the relation between AI benchmarking and intelligence testing, see \cite{chollet:intelli}.
}
The question then cannot be ``to what extent does system $\Sigma$ possess capability $c$'', it has to be ``to what extent can algorithm $A$ learn $c$ from dataset $D$?'' --- and what does that tell us?

\subsection{Single-Task Models}
\label{sec:single}

Let's assume we have defined a task $T$ that we are sufficiently convinced is well represented by dataset $D$. We have trained a model $M$ that performs well on this dataset. What have we learned? We have learned that a learning algorithm of the type of $M$ can model $D$. Further, we have learned that the information to do task $T$ (as exemplified in $D$), is contained in $D$, and $M$ can pick it up.

Under what conditions can we now say that we have modelled $T$, rather than just $D$? If we have convinced ourselves that $D$ represents $T$ faithfully, then we might be willing to make this leap, and with it, claim that we have modelled $C_T$. But we can get further support by collecting more data $D'$ that also exemplifies $T$, but perhaps operationalises it differently. The prediction should be at least that the learning algorithm can also learn to model $D'$; but more significantly, we'd also want the model $M$ learned from $D$ to perform well on $D'$.
Similarly, if we have another task $T'$ of which we think that it involves similar capabilities, we should expect it to be amenable to being 
modelled with a learning algorithm of similar type to $M$.

What do we learn from a model $M'$ (introducing architectural innovation $\kappa$ over $M$) performing better on $T$ (via $D$)? We can take this as indication that $\kappa$ may be what is responsible for increasing performance, and hence what is leading to a more adequate model of $C_T$.

\subsection{Multi-Task Models}
\label{sec:multi}

With the advent of pre-training in NLP \cite{petersetal:elmo,devlinetal:bert}, where a model is trained on (a typically large amount of) data under a specific task-regime (typically language modelling, i.e.\ the task of predicting the next word in a running text) and then becomes part of the model for a target task, it has become common to test on a collection of tasks \cite{Wang2019,superGLUE}.
What do we learn from such a setup? In our Figure~\ref{fig:bench}, if we find a task on which we can pre-train a model $M_P$ that becomes a part of models $M$ and $M'$, and which makes them more powerful than models that do not have access to the pre-trained model, then we can infer that whatever $M_P$ models is a shared part of $M$ and $M'$ as well (and hence \emph{involves} the hypothesised joint capability $C'$). This then provides an instrument to study the tasks: if the pre-trained model works well on some but not all, there must be something that those groups have in common. To make this intelligible, however, recourse to theoretical terms must again be taken. (E.g., assuming that these tasks involve the use of certain types of representation, or certain actions over representations.)


\section{But Are We Making Progress?} 
\label{sec:progress}

Within the logic of this methodology, we are clearly making enormous progress at two links in the chain illustrated in Figure~\ref{fig:bench}: For many of the established tasks, models have been and continue to be proposed that perform better, according to the metrics defined for the tasks. In addition, for many of the tasks, better datasets have been collected, avoiding exploitable biases. Where there is less activity is in systematically studying the implications of success at one task for success at others. The presentation above was largely idealised (or normative): In reality, there is very little explicitness about the assumed connection between tasks and capabilities, and no theory of how (or whether) language competence decomposes into capabilities that could be learned separately and then be assembled into a whole, and there is very little explicit knowledge about the vertical links in the Figure, from one task / model to the next. 

\section{Conclusions}
\label{sec:conc}

In this short paper, we have discussed the methodology of using \emph{language tasks} to drive research on models of language competence. We have argued that the success of this approach hinges on how well progress on one task can be translated into progress on other tasks. While some steps have been taken in this direction, current work still appears to mostly focus on isolated tasks (or groups of tasks). Overcoming this, in our opinion, will require more explicit considerations about how tasks and capabilities are connected, and how the set of capabilities is structured. For this, a (re-)connection with the fields that study the composition of language competence, linguistics and cognitive and developemental psychology, seems to be advisable. Finally, the stark difference between how humans can pick up new tasks, based on an intensional description and a few examples, and how current models do this (by needing massive amounts of data), and between how human language competence develops (from simpler interactions to more complex ones) and how models ``develop'' (from scratch on any task) has to be accounted for, pointing to a future where the real benchmark might be the developmental trajectory, and the tasks are only measures, not targets.

\bibliographystyle{acl_natbib}
\bibliography{/Users/das/work/projects/MyDocuments/BibTeX/all-lit.bib}

\begin{thebibliography}{19}
\expandafter\ifx\csname natexlab\endcsname\relax\def\natexlab#1{#1}\fi

\bibitem[{Antol et~al.(2015)Antol, Agrawal, Lu, Mitchell, Batra, Zitnick, and
  Parikh}]{VQA2015}
Stanislaw Antol, Aishwarya Agrawal, Jiasen Lu, Margaret Mitchell, Dhruv Batra,
  C.~Lawrence Zitnick, and Devi Parikh. 2015.
\newblock Vqa: Visual question answering.
\newblock In \emph{International Conference on Computer Vision (ICCV)}.

\bibitem[{Artstein and Poesio(2008)}]{ArtPoesio:kappa}
Ron Artstein and Massimo Poesio. 2008.
\newblock \href {https://doi.org/10.1136/bmj.312.7039.1166} {{Inter-Coder
  Agreement for Computational Linguistics}}.
\newblock \emph{Computational Linguistic}, 34(4):555--596.

\bibitem[{Borsboom(2005)}]{borsboom:measuring}
Denny Borsboom. 2005.
\newblock \emph{Measuring the Mind: Conceptual Issues in Contemporary
  Psychometrics}.
\newblock Cambridge University Press, Cambridge, UK.

\bibitem[{Butterfield et~al.(2016)Butterfield, Ngondi, and Kerr}]{oxlexcompsci}
Andrew Butterfield, Gerard~Ekembe Ngondi, and Anne Kerr, editors. 2016.
\newblock \emph{A Dictionary of Computer Science}, 7th edition.
\newblock Oxford University Press, Oxford, UK.

\bibitem[{{Chollet}(2019)}]{chollet:intelli}
Fran{\c{c}}ois {Chollet}. 2019.
\newblock \href {http://arxiv.org/abs/1911.01547} {{On the Measure of
  Intelligence}}.
\newblock \emph{arXiv e-prints}, page arXiv:1911.01547.

\bibitem[{Devlin et~al.(2018)Devlin, Chang, Lee, and
  Toutanova}]{devlinetal:bert}
Jacob Devlin, Ming{-}Wei Chang, Kenton Lee, and Kristina Toutanova. 2018.
\newblock \href {http://arxiv.org/abs/1810.04805} {{BERT:} pre-training of deep
  bidirectional transformers for language understanding}.
\newblock \emph{CoRR}, abs/1810.04805.

\bibitem[{Dua and Graff(2019)}]{Dua:2019}
Dheeru Dua and Casey Graff. 2019.
\newblock \href {http://archive.ics.uci.edu/ml} {{UCI} machine learning
  repository}.

\bibitem[{Goyal et~al.(2017)Goyal, Khot, Summers-Stay, Batra, and
  Parikh}]{Goyal2017}
Yash Goyal, Tejas Khot, Douglas Summers-Stay, Dhruv Batra, and Devi Parikh.
  2017.
\newblock \href {http://arxiv.org/abs/1612.00837} {{Making the V in VQA Matter:
  Elevating the Role of Image Understanding in Visual Question Answering}}.
\newblock In \emph{CVPR 2017}.

\bibitem[{Jabri et~al.(2016)Jabri, Joulin, and van~der Maaten}]{Jabri2016}
Allan Jabri, Armand Joulin, and Laurens van~der Maaten. 2016.
\newblock \href {http://arxiv.org/abs/1606.08390} {{Revisiting Visual Question
  Answering Baselines}}.
\newblock In \emph{European Conference on Computer Vision (ECCV)}.

\bibitem[{{Martinez-Plumed} and {Hernandez-Orallo}(2018)}]{aibench}
Fernando {Martinez-Plumed} and Jos{\'e} {Hernandez-Orallo}. 2018.
\newblock \href {https://doi.org/10.1109/TG.2018.2883773} {Dual indicators to
  analyse ai benchmarks: Difficulty, discrimination, ability and generality}.
\newblock \emph{IEEE Transactions on Games}, pages 1--1.

\bibitem[{van Miltenburg(2019)}]{emielphdthesis}
Emiel van Miltenburg. 2019.
\newblock \href
  {https://emielvanmiltenburg.nl/wp-content/uploads/2019/08/phdthesis.pdf}
  {\emph{Pragmatic factors in (automatic) image description}}.
\newblock Ph.D. thesis, Vrije Universiteit Amsterdam.

\bibitem[{Pavlick and Kwiatkowski(2019)}]{Pavlick2019}
Ellie Pavlick and Tom Kwiatkowski. 2019.
\newblock {Inherent Disagreements in Human Textual Inferences}.
\newblock \emph{Transactions of the Association for Computational Linguistics},
  7:677--694.

\bibitem[{Peters et~al.(2018)Peters, Neumann, Iyyer, Gardner, Clark, Lee, and
  Zettlemoyer}]{petersetal:elmo}
Matthew Peters, Mark Neumann, Mohit Iyyer, Matt Gardner, Christopher Clark,
  Kenton Lee, and Luke Zettlemoyer. 2018.
\newblock \href {https://doi.org/10.18653/v1/N18-1202} {Deep contextualized
  word representations}.
\newblock In \emph{Proceedings of the 2018 Conference of the North {A}merican
  Chapter of the Association for Computational Linguistics: Human Language
  Technologies, Volume 1 (Long Papers)}, pages 2227--2237, New Orleans,
  Louisiana. Association for Computational Linguistics.

\bibitem[{Popper(1934)}]{popper:logik}
Karl Popper. 1934.
\newblock \emph{Logik der Forschung}.
\newblock Mohr Siebeck, T{\"u}bingen, Germany.

\bibitem[{Pustejovsky and Stubbs(2013)}]{pustestubbs:annotation}
James Pustejovsky and Amber Stubbs. 2013.
\newblock \emph{Natural Language Annotation for Machine Learning}.
\newblock O'Reilly, Sebastopol, CA, USA.

\bibitem[{Schlangen(2019)}]{schlangen:tasks}
David Schlangen. 2019.
\newblock \href {http://arxiv.org/abs/1908.10747} {Language tasks and language
  games: On methodology in current natural language processing research}.
\newblock \emph{CoRR}, abs/1908.10747.

\bibitem[{Wang et~al.(2019{\natexlab{a}})Wang, Pruksachatkun, Nangia, Singh,
  Michael, Hill, Levy, and Bowman}]{superGLUE}
Alex Wang, Yada Pruksachatkun, Nikita Nangia, Amanpreet Singh, Julian Michael,
  Felix Hill, Omer Levy, and Samuel~R. Bowman. 2019{\natexlab{a}}.
\newblock \href {http://arxiv.org/abs/1905.00537} {{SuperGLUE: A Stickier
  Benchmark for General-Purpose Language Understanding Systems}}.
\newblock In \emph{NeurIPS}, July, pages 1--30.

\bibitem[{Wang et~al.(2019{\natexlab{b}})Wang, Singh, Michael, Hill, Levy, and
  Bowman}]{Wang2019}
Alex Wang, Amanpreet Singh, Julian Michael, Felix Hill, Omer Levy, and
  Samuel~R. Bowman. 2019{\natexlab{b}}.
\newblock \href {http://arxiv.org/abs/1804.07461} {{GLUE: A Multi-Task
  Benchmark and Analysis Platform for Natural Language Understanding}}.
\newblock In \emph{ICLR 2019}, pages 1--20.

\bibitem[{Williams et~al.(2018)Williams, Nangia, and Bowman}]{Williams2018}
Adina Williams, Nikita Nangia, and Samuel~R. Bowman. 2018.
\newblock \href {http://arxiv.org/abs/1704.05426} {{A Broad-Coverage Challenge
  Corpus for Sentence Understanding through Inference}}.
\newblock In \emph{Proceedings of the Conference of the North American Chapter
  of the Association for Computational Linguistics (NAACL)}.

\end{thebibliography}

\end{document}